\documentclass{jlcl}

\newcommand{\authornames}{Katrien Beuls\footnote{Faculté d'informatique, Université de Namur, Belgium} \& Paul Van Eecke\footnote{Artificial Intelligence Laboratory, Vrije Universiteit Brussel, Belgium}}

\newcommand{\authorlastnames}{Beuls \& Van Eecke}

\newcommand{\articletitle}{Construction Grammar and Artificial Intelligence}

\newcommand{\shortarticletitle}{Construction Grammar and Artificial Intelligence}

\usepackage[utf8]{inputenc}
\usepackage[LFE,LAE,T1]{fontenc}
\usepackage[ngerman, main=english]{babel} 

\usepackage{ifxetex}
\usepackage{subcaption}
\usepackage{placeins}
\usepackage{arabtex}

\usepackage{enumitem}
\usepackage{hyperref}
\usepackage{framed}

\ifxetex
\else
\usepackage{CJKutf8}
\fi

\usepackage{natbib}

\title{\articletitle}
\author{AUTHOR NAME\\
AUTHOR INSTITUTION\\
\texttt{AUTHOR EMAIL} \and
AUTHOR 2 NAME\\
AUTHOR 2 INSTITUTION\\
 \texttt{AUTHOR 2 EMAIL}}

\begin{document}

\noindent
\begin{framed}
\noindent
\textcolor{blue}{\textit{Peer-reviewed author's draft of a chapter to appear in the Cambridge Handbook of Construction Grammar (2024 -- edited by Mirjam Fried and Kiki Nikiforidou).}}
\end{framed}

\setarab 
\vocalize 
\transtrue 

\setcounter{page}{1}
\thispagestyle{plain}

\authordata



\begin{abstract}
\noindent
In this chapter, we argue that it is highly beneficial for the contemporary construction grammarian to have a thorough understanding of the strong relationship between the research fields of construction grammar and artificial intelligence.  We start by unravelling the historical links between the two fields, showing that their relationship is rooted in a common attitude towards human communication and language.  We then discuss the first direction of influence,  focussing in particular on how insights and techniques from the field of artificial intelligence play an important role in operationalising, validating and scaling constructionist approaches to language.  We then proceed to the second direction of influence,  highlighting the relevance of construction grammar insights and analyses to the artificial intelligence endeavour of building truly intelligent agents. We support our case with a variety of illustrative examples and conclude that the further elaboration of this relationship will play a key role in shaping the future of the field of construction grammar.
\end{abstract}

\section{A common attitude towards communication and language}

To many contemporary linguists, construction grammar (CxG) and artificial intelligence (AI) might not spring to mind as two scientific disciplines that are closely related.  Yet, both fields share a long-standing interest in modelling human communication and language, and adopt a similar attitude towards this area of research.  This similar attitude most visibly encompasses the following aspects:

\begin{itemize}
\item \textbf{Language serves a communicative purpose.} The basic function of language is to serve the communicative needs of its users, facilitating the transfer of information from one language user to another.  As such, language production corresponds to the process of expressing an idea in the form of a natural language utterance, while language comprehension corresponds to the process of reconstructing the communicative intention underlying an observed utterance.  

\item \textbf{Communication is a bidirectional process.} Adequate representations and processing mechanisms for linguistic knowledge need to support the bidirectional nature of human communication and language.  This entails that language comprehension and production are performed using the same representations and processing mechanisms.  It is crucial for both humans and artificial agents that they can use the linguistic knowledge they have acquired through language comprehension in the production direction, and that they can understand themselves any utterances they produce.  

\item \textbf{Languages are acquired rather than innate.} An individual language user acquires the language of their community by actively taking part in situated, communicative interactions.  Languages cannot be innate,  as this would compromise their ability to dynamically adapt to changes in the environment or in the communicative needs of their users.  As language processing is heavily intertwined with other cognitive processes,  in particular reasoning and vision, it is preferably modelled through the same general cognitive mechanisms.

\item \textbf{Languages emerge and evolve through communication.} Each individual language user has built up their own linguistic system based on the communicative interactions they have participated in.  This linguistic system is unique to each language user,  as it has been shaped by the history of their successes and failures in communication.  The evolutionary processes of variation and selection that take place in each individual during  communication ensure that the linguistic system of each individual is compatible on a communicative level with the linguistic systems of all other individuals in the population.

\item  \textbf{Languages are grounded in (knowledge of) the world.} As the basic function of language is to serve the communicative needs of its users,  it is necessarily grounded in the world in which they live.   Understanding and producing natural language expressions heavily relies on world knowledge and common-sense reasoning.  Indeed,  the intended meaning underlying a natural language expression crucially depends on the concrete situation in which it was uttered.  This situation includes a variety of aspects, including observed objects and actions in the world,  pragmatic and discursive factors,  and inter-personal relations.
\end{itemize}

It is clear that the research fields of construction grammar and artificial intelligence adopt a similar attitude towards the study of human language and communication. Especially in the period from the late sixties to the early nineties, this could be witnessed by collaborations and close interactions  between leading figures in both fields. Today, traces of these interactions are still visible through a close reading of contemporary articles and textbooks.  For example, Charles Fillmore, who is often considered the founding father of the field of construction grammar, explicitly acknowledges in his seminal paper on the case of let alone \citep{fillmore1988regularity} the advice of UC Berkeley AI professor Robert Wilensky and his student Peter Norvig, who later went on to become a key figure in AI education worldwide \citep{russell2021artificial}. The advice was bidirectional, as Fillmore had served as a member of the PhD jury of Peter Norvig in 1978.  Fillmore's case grammar \citep{fillmore1968case} had a substantial influence on later natural language understanding systems and was even presented as a standard component of natural language understanding in the first edition of Patrick Winston's standard textbook on artificial intelligence \citep{winston1977artificial,jurafsky2014charles}.  Starting in the mid-seventies,  the notion of a `frame' as a situational representation emerged through an interdiscplinary dialogue between linguists \citep{fillmore1976frame},  sociologists \citep{goffman1974frame},  psychologists \citep{rumelhart1080schemata} and artificial intelligence researchers \citep{minsky1974framework,schank1977scripts}.  Fillmore's linguistic work on frame semantics has thereby been highly influential in the field of artificial intelligence, most notably through the eventual development of the FrameNet project \citep{baker1998berkeley,fillmore2001frame}\footnote{We refer the interested reader to Chapters 2 and 3 of this volume for more background on frame semantics and FrameNets respectively. }. While these examples are anecdotes rather than evidence,  they do reflect that the idea that researchers in construction grammar and artificial intelligence are working towards a common goal, namely to understand and model human language use, was strongly present in the early days of construction grammar. 

When taking a closer look at recent contributions to journals and conferences in construction grammar and artificial intelligence, one gets the impression that interactions between both fields are much scarcer today than they used to be in the past. At the same time, the knowledge that both fields used to be aware of their common ground seems to have vanished to a large extent from both communities. We can only speculate about the reasons for this divergence, and there is probably not a single cause for this effect. Perhaps it is a symptom of a broader tendency of research fields to specialise and isolate. Or it may be a consequence of the reaction of many cognitively inspired linguists against the dogmas of generative grammar, by which they have sometimes overreacted and thereby developed an aversion towards any form of formalisation.  In practice,  informal theories of construction grammar are less attractive to artificial intelligence researchers,  as these researchers often lack the extensive construction grammar expertise that is needed to formalise them.  It could also be due to the progressive institutionalisation of artificial intelligence research groups within computer science departments, by which fewer and fewer linguists are hired and by which research in AI focuses increasingly more on statistics and data science at the expense of models involving domain knowledge.  

It is an explicit goal of this chapter to draw renewed attention to the common goals and similar attitude towards language and communication that have motivated mutually beneficial collaborations between construction grammarians and AI scholars in the past,  and to emphasise the great value that lies in the further elaboration of this relationship.  On the one hand, we focus on the influence of ideas and techniques from the field of artificial intelligence on the field of construction grammar,  thereby discussing the importance of these techniques for operationalising the basic tenets of construction grammar,  for validating the consistency and preciseness of construction grammar theories, for corroborating these theories with corpus data and for scaling constructionist approaches to language.  On the other hand, we zoom in on the importance and use of construction grammar insights and analyses in the field of artificial intelligence,  thereby emphasising the excellent fit between the foundational ideas underlying constructionist approaches to language and the needs of researchers aiming to build truly intelligent systems.  We are convinced that a thorough understanding of the relationship between both fields is highly beneficial for the contemporary construction grammarian,  and that further developments in this direction will play a key role in shaping the future of the field of construction grammar.

This chapter focusses explicitly on approaches within the fields of construction grammar and artificial intelligence that explicitly model constructional language processing.  The relationship between construction grammar and transformer-based large language models is covered in Chapter 22 of this volume.

\section{Artificial intelligence for operationalising construction grammar}

This section discusses how methods and techniques from the field of artificial intelligence have contributed to the formalisation and computational operationalisation of constructionist approaches to language.  It first revisits the basic tenets of construction grammar and then continues with a stepwise explanation of how these basic tenets can be mapped to data structures and algorithms that are known from the field of artificial intelligence.

\subsection{The basic tenets of construction grammar}

Construction grammar refers to a family of linguistic theories that share a number of foundational principles.  These principles,  as laid out by among others  \cite{fillmore1988mechanisms}, \cite{goldberg1995constructions}, \cite{kay1999grammatical}, \cite{croft2001radical}, \cite{fried2004construction} and \cite{hilpert2014construction}, are the following:

\begin{itemize}

\item \textbf{All linguistic knowledge is captured in constructions}. All linguistic knowledge that is needed for language comprehension and production can be represented in the form of form-meaning mappings, called constructions.  These constructions can freely combine to comprehend and produce utterances, as long as no conflicts occur \citep{goldberg2006constructions,vaneecke2018creative}.

\item \textbf{There exists a lexicon-grammar continuum}.  Construction grammars do not distinguish between the traditional notions of ``words'' and ``grammar rules''.  Constructions can range from fully instantiated form-meaning mappings, as in the case of idioms, to abstract schemata, as in the case of argument structure or information structure constructions.  Many constructions are partially instantiated and partially abstract,  as exemplified by  the famous \textit{let-alone} construction \citep{fillmore1988regularity}.

\item \textbf{Constructions span all levels of linguistic analysis}. Constructions can include information from all levels of traditional linguistic analysis.  The form side of a construction typically contains a combination of phonetic, phonological, lexical, morpho-syntactic and multi-modal information, while its meaning side typically combines semantic and pragmatic information.  Constructions do not need to contain information on each of these levels.  For example, they can, but do not need to, include word order constraints.

\item \textbf{Constructions grammars are dynamic systems}.  Constructions are not innate, but \textit{constructed} during communicative interactions.  Based on the frequency of their success and failure in communication,  constructions can become more or less entrenched.  As a consequence,  a construction grammar always represents the linguistic knowledge of an individual language user, as opposed to modelling an imaginary \textit{ideal language user}.    

\item \textbf{Construction grammars should account for all linguistic phenomena}.  Construction grammars do not adhere to the generative core-periphery distinction, and all linguistic phenomena are considered to be of equal interest.  The same machinery is used to handle all linguistic phenomena, whether they are traditionally seen as regular,  semi-regular,  irregular or idiomatic. 

\end{itemize}

Formalisation was considered to be an important aspect of construction grammar research since the inception of the field,  with initial formalisations being inspired by phrase structure grammars \citep[see e.g.][]{fillmore1988mechanisms}.  However, the focus on formalisation faded into the background when the Lakovian/Goldbergian branch of construction grammar,  called \textit{cognitive construction grammar} became predominant.  The focus was on the conceptual clarification of the refreshing ideas that laid the foundations of the field, rather than on precise formalisations or computational implementations.  However,  once the initial ideas had settled,  a relatively small number of construction grammarians started to focus on how these ideas could be formalised, verified,  implemented and tested \citep{kay1999grammatical,steels2004constructivist,bergen2005embodied,feldman2009embodied,sag2012sign,michaelis2013sign}.  Traditional techniques that were commonly used to formalise and implement generative grammars,  such as the unification of feature structures, had to be complemented with innovative machinery that could accommodate those aspects of constructions that substantially differ from phrase structure rules. These include among others the fact that constructions can be non-local,  that they do not necessarily correspond to tree-building operations \citep{vantrijp2016chopping},  that they can, but do not need to, include word order constraints, and that they are acquired through communicative interactions.  

The innovative machinery that was needed to formalise and implement the basic tenets of construction grammar was borrowed from the field of artificial intelligence. In particular,  heuristic search strategies \citep{wellens2010priming,bleys2011search} and innovative unification algorithms \citep{debeule2006unify,sierra2012logic} were used to operationalise the free combination of constructions.  Multi-agent simulations \citep{steels2005emergence,vantrijp2008emergence,beuls2013agent,nevens2022language} were used to model the dynamic nature of construction grammars, including on the one hand the constructivist emergence, evolution and acquisition of constructions, and on the other hand the entrenchment processes that take place in the constructicon.

\subsection{Towards computational construction grammar}

The basic function of language is to support communication,  i.e.  the transfer of information from one language user to another.  There are always two parties involved in a communicative interaction,  namely a party that produces a linguistic expression and a party that comprehends it.  Language production amounts to expressing an idea or intention in the form of a natural language utterance, while language comprehension consists in reconstructing the idea or intention underlying an observed utterance.  Language processing can as a consequence be seen as a bidirectional process of mapping between intentions or ideas, referred to as \textit{meaning}, and natural language utterances that express them, referred to as \textit{form}.  In computational terms, this means that we need to (i) represent natural language utterances, (ii) represent semantic structures, and (iii) provide a model that maps between these representations both in the comprehension and the production direction.

In essence, computationally operationalising construction grammar, or any linguistic theory for that matter, involves finding precise representations and processing mechanisms that correspond to all aspects of the underlying theory. In computational terms, representations take the form of data structures, while processing mechanisms take the form of algorithms that operate over these data structures. When designing and implementing data structures and algorithms that operationalise computational construction grammar, the basic tenets of construction grammar, as laid out in the previous section, serve as a logical starting point. 

In the next sections, we will illustrate how the basic tenets of construction grammar can be captured and operationalised in the form of a computational construction grammar system, highlighting the important role of techniques and methods from the field of artificial intelligence in this endeavour. We will adopt the terminology and conceptual framework underlying Fluid Construction Grammar (FCG -- \url{www.fcg-net.org}) \citep{steels2004constructivist, vantrijp2022fcg,beuls2023fluid}. FCG is a computational construction grammar implementation that takes the form of a special-purpose programming language for designing and computationally implementing construction grammar models.  FCG has the explicit goal of providing computational counterparts to the basic tenets of construction grammar in the form of a library of ready-to-use building blocks, while remaining an open framework that provides the flexibility and customisability to explore novel construction grammar ideas.  For a technical introduction to FCG,  we refer the interested reader to Chapter 3 of \cite{vaneecke2018generalisation}.

\subsubsection{Representing utterances and meanings}

There are many different ways in which natural language utterances and semantic representations can be computationally represented.  For didactic reasons,  we will adopt the representations that are most commonly used in the computational construction grammar literature.  Utterances are represented as a combination of tokens and adjacency constraints between those tokens.  A token is a sequence of characters,  i.e.  a string,  which corresponds to a part of an utterance that is enclosed by white space or punctuation.  In order to be able to unambiguously refer to a token,  each token is assigned a unique identifier.  Adjacency constraints use these unique identifiers to express that two tokens are adjacent to each other.  The tokens and adjacency constraints can be expressed as predicates and an entire utterance can consequently be represented as a set of predicates.  An example of such a representation for the utterance ``\textit{The more you think about it, the less it makes sense.}'' \citep[example from][]{hilpert2021what} is shown here:

\begin{small}
\begin{verbatim}
{string(the-1, "The"), string(more-1, "more"), string(you-1, "you"), 
 string(think-1, "think"), string(about-1, "about"), string(it-1, "it"), 
 string(,-1, ","), string(the-2, "the"), string(less-1, "less"), 
 string(it-2, "it"), string(makes-1, "makes"), string(sense-1, "sense"), 
 string(.-1, "."), adjacent(the-1,  more-1), adjacent(more-1, you-1), 
 adjacent(you-1, think-1), adjacent(think-1, about-1), adjacent(about-1, it-1), 
 adjacent(it-1, ,-1), adjacent(,-1, the-2), adjacent(the-2, less-1), 
 adjacent(less-1, it-2), adjacent(it-2, makes-1), adjacent(makes-1, sense-1), 
 adjacent(sense-1, .-1)}
\end{verbatim}

\end{small}

This representation consists of 13 `string' predicates that represent the tokens in the utterance along with their unique identifiers,  and 12 `adjacent' predicates that encode the order of the tokens within the utterance.

We represent semantic structures using the Abstract Meaning Representation (AMR) formalism \citep{banarescu2013abstract}.  AMR is a meaning representation language that was developed for representing the meaning of utterances in a way that (i) abstracts away from syntactic idiosyncrasies, (ii) is easy to read for humans, and (iii) is easy to manipulate by computers \citep{banarescu2013abstract}.  An example of the AMR representation for the utterance ``\textit{The more you think about it, the less it makes sense.}'' introduced above is shown here:  

\begin{small}
\begin{verbatim}
(c / correlate-91
   :ARG1 (m / more
            :ARG3-OF (h / have-degree-91
                        :ARG1 (t / think-01
                                 :ARG0 (y / you)
                                 :ARG1 (i / it))))
   :ARG2 (l / less
            :ARG3-OF (h2 / have-degree-91
                        :ARG1 (s / sense-02
                                 :ARG1 i))))
\end{verbatim}
\end{small}

On the highest level,  we can observe that the utterance evokes a correlation, introduced by the \texttt{correlate-91} roleset \citep{bonial2018abstract}.  By definition, this roleset describes the correlation between two degrees to which two things hold.  The first degree,  notated as \texttt{:arg1},  is a relation, in this case \texttt{more}.  This relation correlates with the second degree,  notated as \texttt{:arg2}, in this case the relation \texttt{less}.  Thus,  an increase in something leads to a decrease in something else.  The first relation, \texttt{more},  corresponds itself to the degree (\texttt{:arg3-of} of the \texttt{have-degree-91} roleset) to which a thinking event of roleset \texttt{think-01} holds.  The agent/thinker (\texttt{:arg0}) of the thinking event is \texttt{you} while the undergoer/thought (\texttt{:arg1}) of the thinking event is \texttt{it}.  Thus,  the \texttt{more} relation embodies the \texttt{degree} to which \texttt{you} think about \texttt{it}.  The second relation, \texttt{less}, corresponds to the degree (\texttt{:arg3-of} of the \texttt{have-degree-91} roleset) to which a sense-making event of roleset \texttt{sense-02} holds.  The `thing that makes sense' (\texttt{:arg1}) is the same entity (\texttt{i}) as the undergoer/thought of the thinking event,  namely \texttt{it}. The \texttt{less} relation embodies thus the  \texttt{degree} to which \texttt{it} makes sense, with \texttt{it} being the thing you are thinking about.  In sum, the AMR representation of the utterance ``\textit{The more you think about it, the less it makes sense.}'' expresses that there is a correlation between the increasing degree to which you think about a particular referent and the decreasing degree to which that referent makes sense.

The AMR representation shown above is expressed using the Penman notation, which was designed to be maximally human-readable.  For computational purposes, we use a different notation,  which represents AMR structures in the form of sets of predicates.  As a consequence,  meaning representations can be represented using the same data structure as linguistic utterances. The translation from Penman notation to sets of predicates is lossless and automatic.  The corresponding set of predicates for the example above is the following:

\begin{small}
\begin{verbatim}
{correlate-91(c), more(m), have-degree-91(h), think-01(t), you(y), it(i), 
 less(l), have-degree-91(h2), sense-02(s), :arg1(c, m), :arg2(c, l), 
 :arg3-of(m, h), :arg1(h, t), :arg0(t, y), :arg1(t, i), :arg3-of(l, h2), 
 :arg1(h2, s), :arg1(s, i)}
\end{verbatim}
\end{small}

\subsubsection{Language comprehension and production}

Now that we have established representations for utterances and semantic structures,  we can define language comprehension and production as processes that map between these representations.  In computational construction grammar,  the linguistic knowledge that drives these processes is captured in the form of constructions.  Intuitively,  the task of these constructions is to move from a representation of an utterance to a representation of its meaning and vice versa.  

In order to operationalise constructional language processing,  computational construction grammar frames language processing as a search problem.  Search problems form a class of problems that have been extensively studied in artificial intelligence, and of which the foundations date back to the seminal work of \cite{newell1956logic}.  Search problems are characterised by three main components: (i) a representation of the state of the search problem, (ii) operators that can work on a problem state and create new problem states that are hopefully closer to a solution, and (iii) a goal test that determines whether a problem state corresponds to a solution or not.  In the case of constructional language processing, these three components are instantiated as follows \citep{vaneecke2017metalayer}:

\begin{enumerate}
\item \textit{Transient structures} serve as the representation of the state of the search problem.  A transient structure holds all information that is known about an utterance being comprehended or produced at a given point during processing.
\item \textit{Constructions} serve as the operators of the search problem.  Given a transient structure,  they can contribute new information and thereby give rise to a new transient structure.
\item \textit{Goal tests} verify whether a given transient structure qualifies as a solution to the search problem.  
\end{enumerate}

\begin{figure}
\centering
\begin{subfigure}[b]{.35\textwidth}
\includegraphics[width = \textwidth]{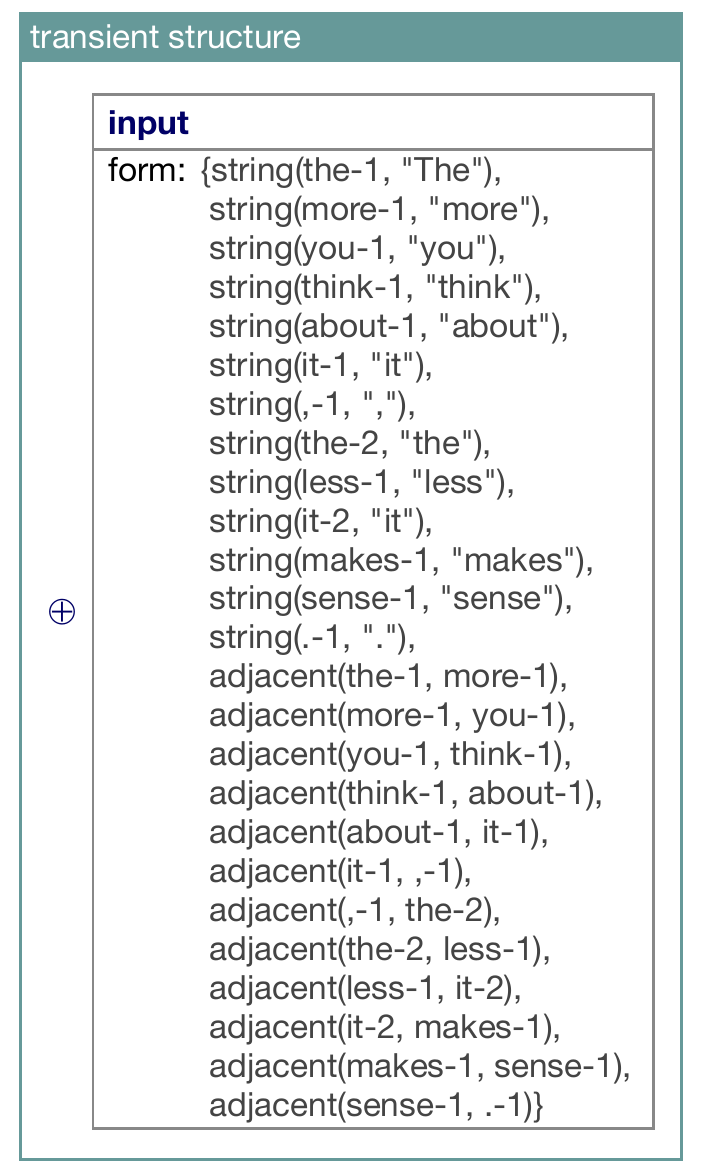}
\caption{Comprehension}
\end{subfigure}
\begin{subfigure}[b]{.315\textwidth}
\includegraphics[width = \textwidth]{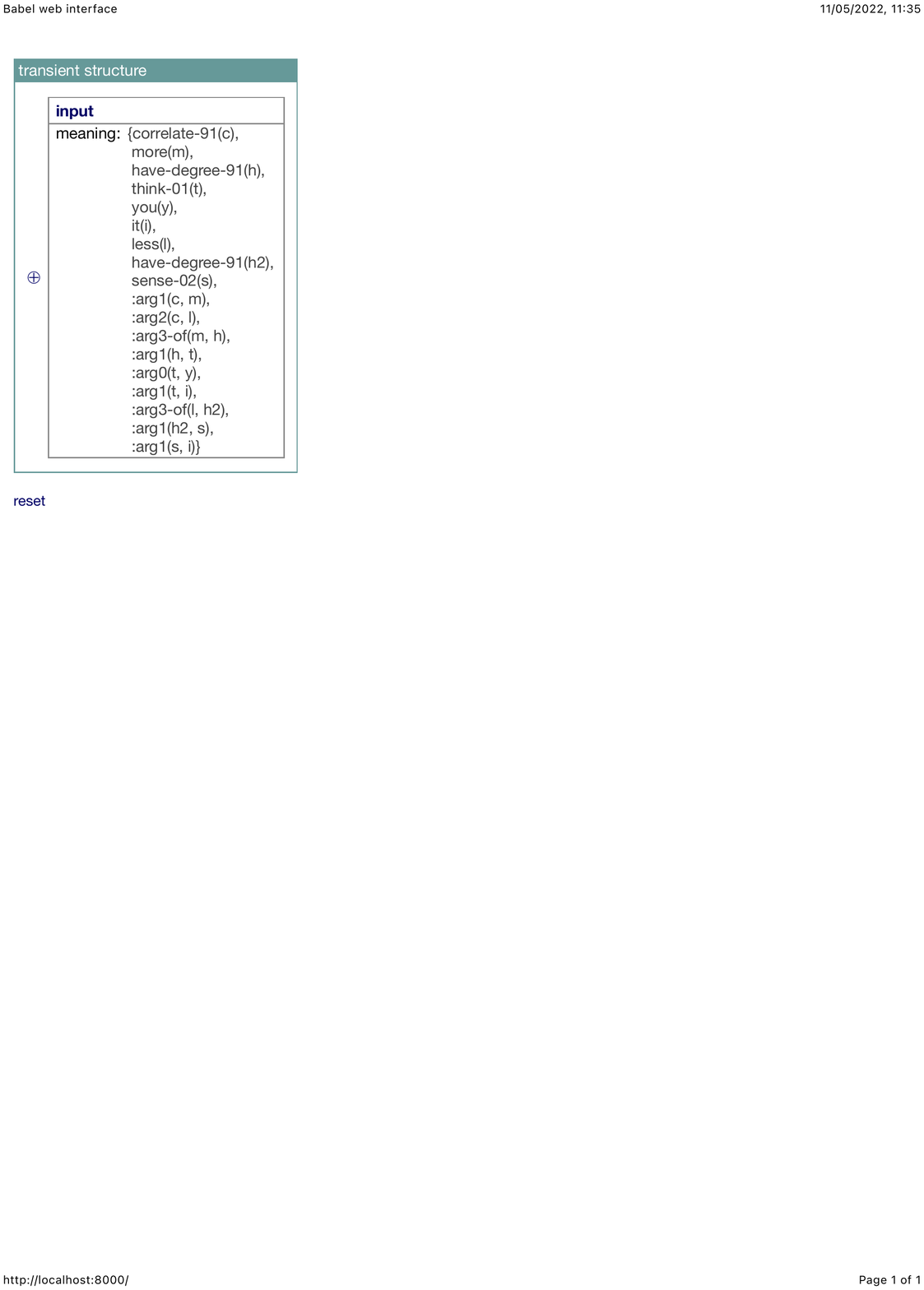}
\caption{Production}
\end{subfigure}
\caption{Initial transient structures in comprehension (a) and production (b) for the utterance ``\textit{The more you think about it, the less it makes sense.}''}
\label{fig:initial-ts}
\end{figure}

The search process starts from a representation of the problem to be solved.  In the case of constructional language processing,  this representation takes the form of an \textit{initial transient structure}.  By definition, the initial transient structure
holds all information that is known before processing starts.  In the comprehension direction, the initial transient structure contains the form to be comprehended.  In our example, this corresponds to the set of string and adjacency predicates introduced above. In the production direction, the initial transient structure contains a representation of the meaning to be expressed.  In our example, this is the AMR representation shown above.  The initial transient structures for comprehending and producing the example utterance, namely ``\textit{The more you think about it, the less it makes sense.}'', are shown Figure \ref{fig:initial-ts}.  The initial transient structures store their information in an `input' unit under a `form' and `meaning' feature respectively,  denoting that it was the initial input to the problem solving process.  

Constructions capture linguistic information that can be used to advance the comprehension and production problem solving processes.  Given a transient structure, a construction can contribute new linguistic information and thereby create a new transient structure, which is hopefully closer to a solution.  Constructions consist of two parts,  a conditional pole and a contributing pole.  The conditional pole contains the preconditions for a construction to apply,  and thereby create a new transient structure  through its application.  The contributing pole contains the postconditions of the construction, i.e. information that will be added to the new transient structure during the application of the construction.  As constructions support both the comprehension and production of utterances, they hold two sets of preconditions,  one for comprehension and the other for production.  Preconditions in the comprehension direction serve as additional postconditions in the production direction and vice versa. During constructional language processing, constructions check whether their preconditions are compatible with a given transient structure in a given direction of processing,  and if this is the case,  they create a new transient structure that extends the current transient structure with the information contained in their contributing pole, combined with their preconditions of the other direction of processing.

\begin{figure}
\centering
\includegraphics[width=.7\textwidth]{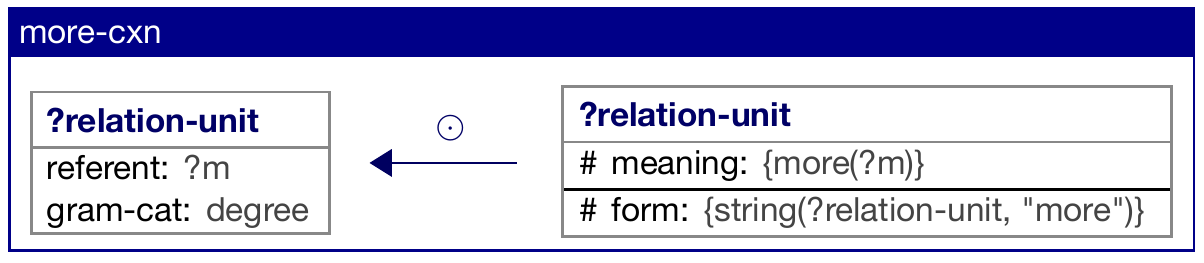}
\caption{The \textit{more-cxn},  which maps between the form ``more'' and the meaning predicate \textit{more(?m)},  contributing the information that the result is of grammatical category `degree' and that the referent of the unit is the argument of the \textit{more} predicate.}
\label{fig:more-cxn}
\end{figure}

\begin{figure}
\centering
\begin{subfigure}[b]{.35\textwidth}
\includegraphics[width = \textwidth]{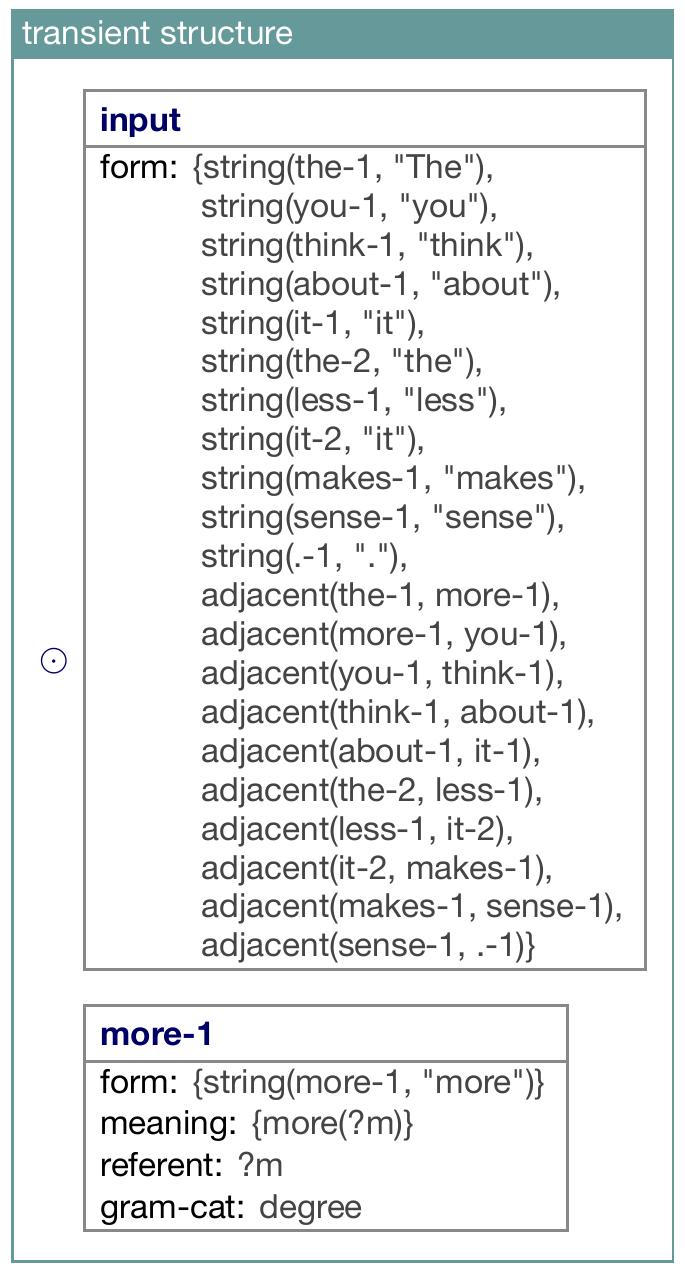}
\caption{Comprehension}
\end{subfigure}
\begin{subfigure}[b]{.315\textwidth}
\includegraphics[width = \textwidth]{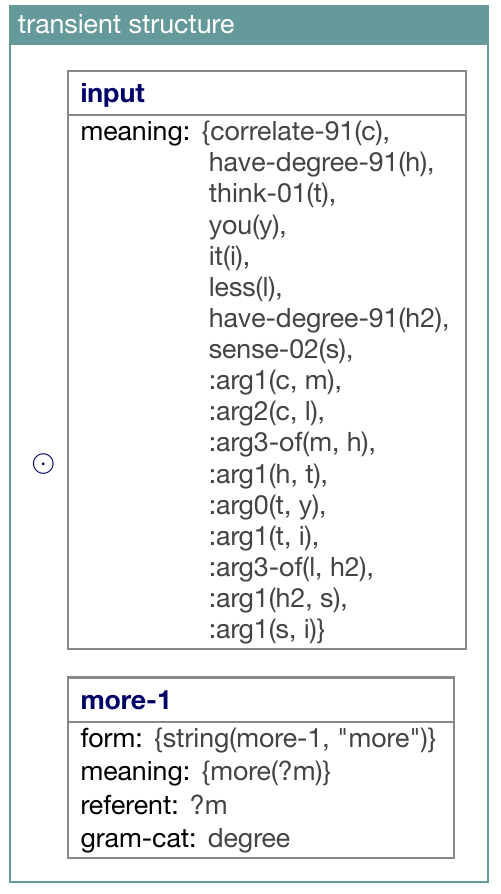}
\caption{Production}
\end{subfigure}
\caption{Transient structures in comprehension (a) and production (b) after applying the \textsc{more-cxn} from Figure \ref{fig:more-cxn} to the initial transient structures shown in Figure \ref{fig:initial-ts}.}
\label{fig:ts-after-more}
\end{figure}

An example of a construction is shown in Figure \ref{fig:more-cxn}.  The name of the construction, here \textsc{more-cxn}, is written in the blue box. The preconditions of the construction are written on the right-hand side of the horizontal arrow, while the postconditions are written on its left-hand side.  The preconditions for comprehension and production are separated by a horizontal line, with the preconditions for production being written above the line and those for comprehension below it.  On a conceptual level,  the \textsc{more-cxn}  maps between the form ``more'' and the AMR predicate \textit{more(?m)}, contributing the information that the resulting unit is of grammatical category `degree' and that its referent is the argument of the \textit{more} predicate.  Technically,  it does this through two features on its conditional pole and two features on its contributing pole.  On its conditional pole,  the construction contains a precondition for comprehension that a form predicate \textit{string(?relation-unit, ``more'')} should be part of the input,  as well as a precondition for production that a meaning predicate \textit{more(?m)} should be part of the input\footnote{In fact, the \# sign that precedes these features explicitly indicates that they should be found in the `input' unit rather than in any unit of the transient structure.}.  If this is the case,  the construction can apply and a new transient structure is created.  This new transient structure starts as a copy of the current transient structure.  In comprehension, the information is added that a meaning predicate \textit{more(?m)} is involved, along with the information that this unit is of grammatical category `degree' and that its referent is the argument of the \textit{more} predicate. In production, the information is added that a string ``more'' is involved, along with the information that the unit is of grammatical category `degree' and that its referent is the argument of the \textit{more} predicate.  The result of the application of the \textsc{more-cxn} shown in Figure \ref{fig:more-cxn} on the transient structures shown in Figure \ref{fig:initial-ts} is provided in Figure \ref{fig:ts-after-more}.

Every time a new transient structure has been created as the result of a successful construction application, a number of \textit{goal tests} are automatically run on this new transient structure to verify whether it qualifies as a solution state \citep{bleys2011search}. Typical goal tests for constructional language processing include (i) checking whether no more constructions can apply, (ii) verifying whether all `string' or `meaning' predicates have been processed,  and (iii) checking whether the meaning comprehended so far consists of a fully connected network of predicates linked through their arguments.  As soon as all goal tests succeed for a given transient structure, it is flagged as a solution state and the search process is halted.  Depending on the direction of processing,  all predicates under a `meaning' (comprehension) or `form' (production) feature are extracted from the solution transient structure.  The result of the comprehension process is  a set of meaning predicates, while the result of the production process is a set of string and adjacency predicates which can be automatically rendered as an utterance.

As it is typically the case that multiple constructions can apply to a given transient structure, the search space involved in the exploration of alternative construction applications quickly grows very large. To navigate the search space in an informed way, researchers in AI have developed \textit{heuristic search} techniques.  These techniques rank problem states according to their quality,  estimating for example how close they are to a solution state \citep{pearl1984heuristics,russell2021artificial}.  Common heuristics for steering the search space involved in constructional language processing include the number of constructions applied so far (favouring deeper solutions) and the number of units that were matched during construction application (favouring constructions that span larger patterns).  More recently,  it has been shown that neural sequence-to-sequence-based heuristics perform particularly well at ranking transient structures based on the sequence of constructions that have been applied in order to reach them \citep{vaneecke2022neural}.  

\begin{figure}
\centering
\includegraphics[width=\textwidth]{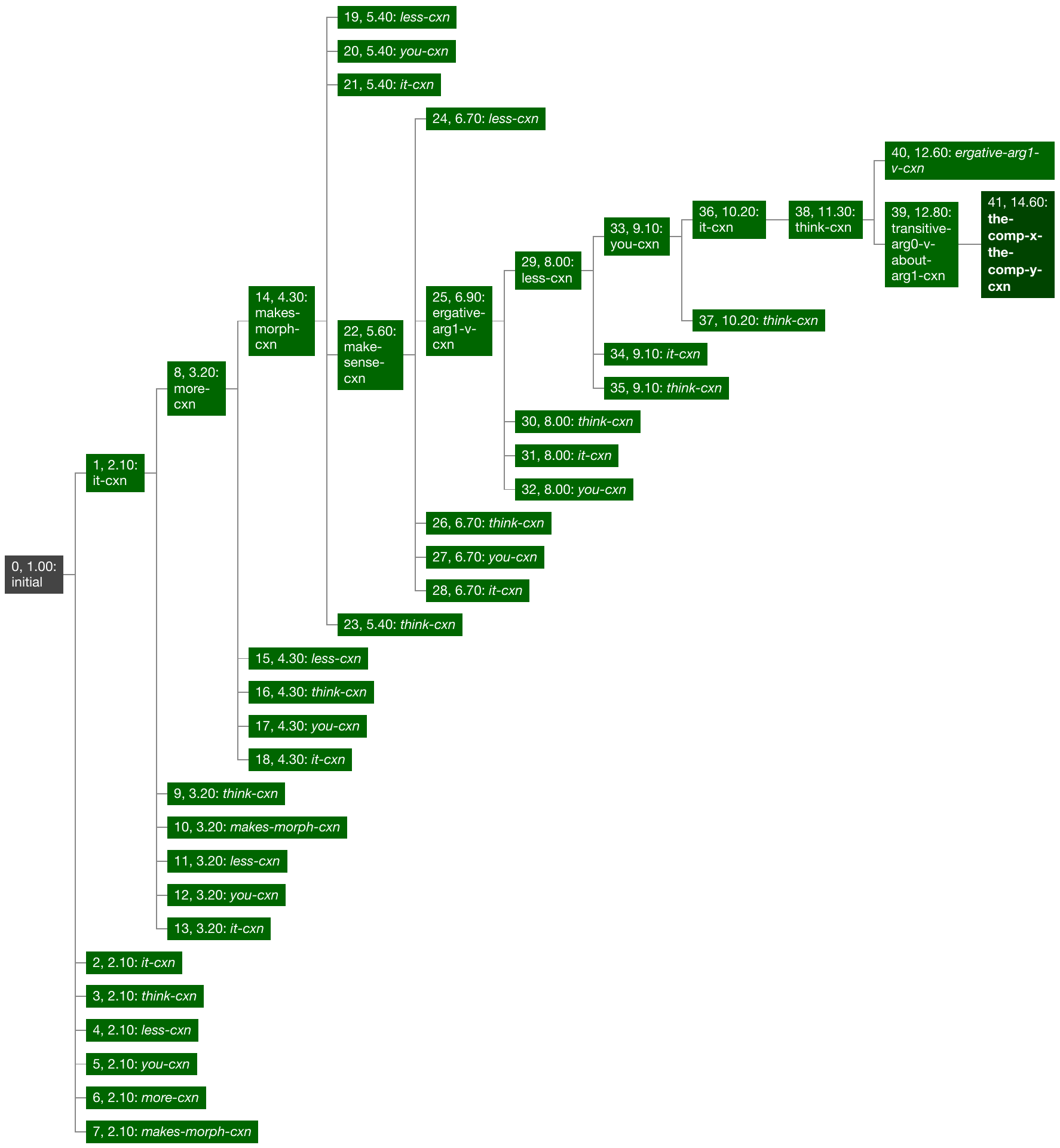}
\caption{The search space involved in the comprehension of the utterance ``\textit{The more you think about it, the less it makes sense.}''. A solution is found in node 41, after the application of 11 constructions.}
\label{fig:search}
\end{figure}

\begin{figure}
\centering
\includegraphics[width=.75\textwidth]{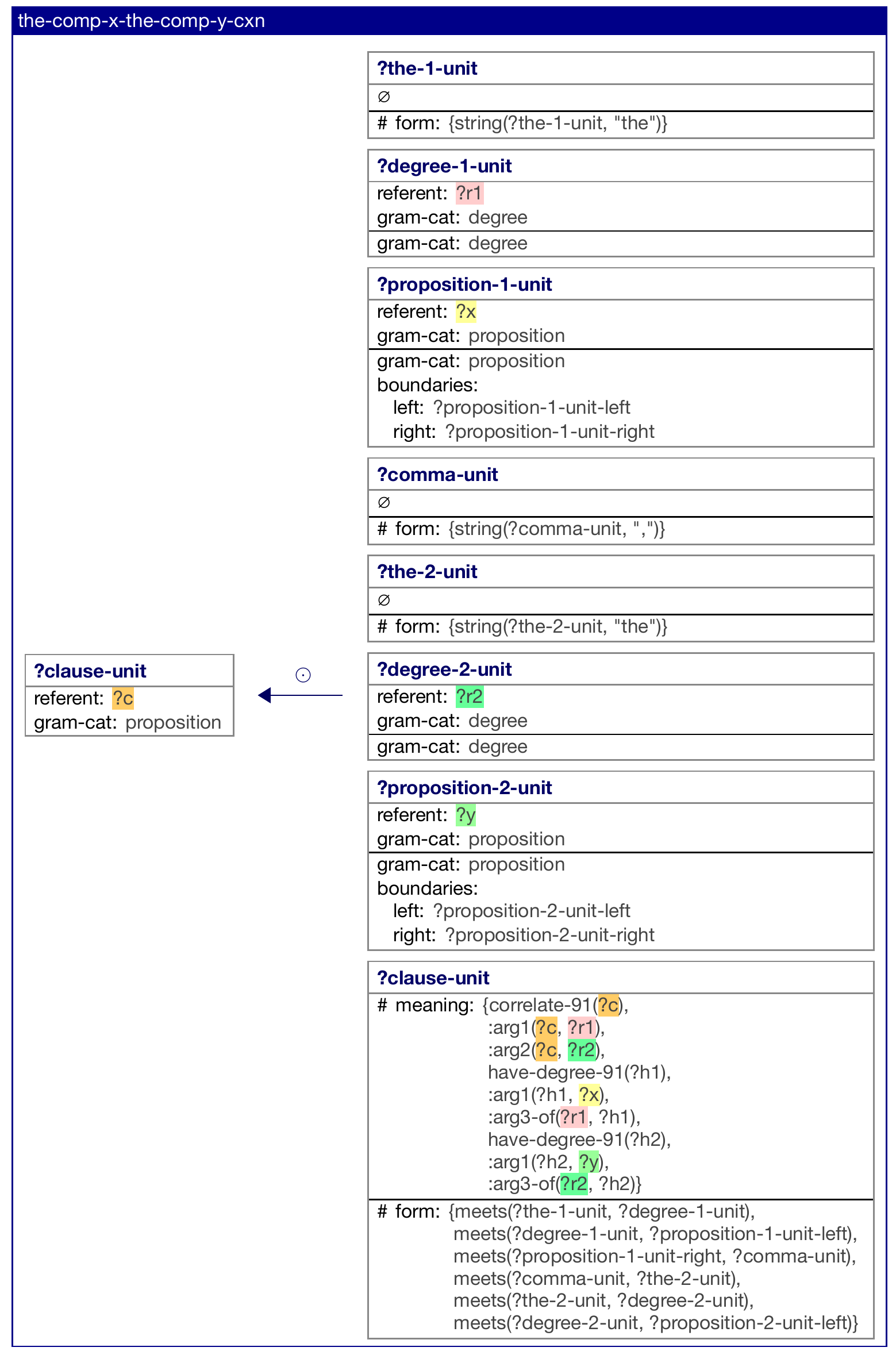}
\caption{The \textsc{the-comp-X-the-comp-Y-cxn} pairs the pattern [``the''-\textit{degree}-\textit{proposition}-``,''-``the''-\textit{degree}-\textit{proposition}] with its meaning representation that states that the extent to which the first degree holds for the first proposition is correlated to the extent to which the second degree holds for the second proposition.}
\label{fig:comparative-cxn}
\end{figure}

Figure \ref{fig:search} shows the search space involved in the comprehension process of the example utterance ``\textit{The more you think about it, the less it makes sense}''.  The initial transient structure is shown in grey in the left of the Figure.  The branching tree that is drawn to the right represents all construction applications that have taken place.  The resulting transient structures are numbered according to when they were created (first number) and explored in the order obtained through their heuristic value (second number).  In this example,  all goal tests succeed for transient structure 41 shown in dark green.  This transient structure is the result of the 11 construction applications that led from the initial transient structure to the solution transient structure.  The last construction that applied was the \textsc{the-comp-x-the-comp-y-cxn}, a high-level construction that pairs the pattern [``the''-\textit{degree}-\textit{proposition}-``,''-``the''-\textit{degree}-\textit{proposition}] with its meaning representation that states that the extent to which the first degree holds for the first proposition is correlated with the extent to which the second degree holds for the second proposition.  An implementation of this construction in FCG is shown in Figure \ref{fig:comparative-cxn}.  The construction includes adjacency constraints that capture the word order inherent to the pattern, as well as meaning predicates that integrate the referents of the different components of the pattern.

\subsubsection{Acquisition, evolution and entrenchment}
\label{sec:games}

As construction grammars are dynamic systems that are \textit{constructed} during communicative interactions, the inventory of features and categories that is used in an individual grammar is open-ended. In Fluid Construction Grammar, this is reflected by the absence of an a priori specification of possible features and their values. The fact that new features and values can be dynamically added should the need arise, is a necessary precondition for modelling the invention, adoption and evolution of constructions in the context of both language acquisition and language emergence. In such experiments, inspiration is again drawn from the field of artificial intelligence, this time from research on learning in multi-agent systems. These experiments typically consist in a community of language users being modelled as a population of autonomous agents that participate in pairwise, goal-driven communicative interactions, referred to as \textit{language games} \citep{steels1998origins,steels2001language}. 

Language games either adopt a tutor-learner scenario or a language emergence scenario. In a tutor-learner scenario, the goal is that one or more learner agents acquire the language of the community in a constructivist manner. In an emergence scenario, an entirely new language emerges that satisfies the communicative needs of its members. 
A typical language game in an emergence scenario proceeds as follows. At the beginning of each interaction, two agents are selected from the population and are assigned the role of either speaker or hearer. The agents are placed in a particular scene and need to successfully communicate to solve a given task, for example referring to objects or events that they observe in the scene. The agents are equipped with mechanisms for inventing and adopting linguistic means (i.e. constructions) that can be needed to achieve communicative success. After each interaction, the speaker provides feedback to the hearer about the outcome of the task. This allows the hearer to learn in the case that the agents did not reach communicative success. Additionally, both agents reward the constructions that were used in the case of a successful interaction, and punish them in the case of a failed interaction. As more and more interactions take place, the agents in the population gradually converge on a shared language \citep{devylder2006reach}. The language of each individual agent has been shaped by the communicative interactions it has participated in and is, therefore, well adapted to the task and the environment. As the scores of individual constructions reflect their frequency of successful application, they can be seen as a measure of their degree of \textit{entrenchment}, and are as a consequence often referred to by the term \textit{entrenchment scores}. When comprehending and producing linguistic utterances, the entrenchment scores are used to prioritise constructions where multiple constructions are in competition with each other. Notable applications of this language acquisition and emergence paradigm include experiments on the emergence and evolution of phonetic systems \citep{deboer2000self, oudeyer2006self}, vocabularies \citep{baronchelli2006sharp, steels2015talking}, domain-specific conceptual systems \citep{steels2005coordinating, bleys2009linguistic, spranger2016evolution, nevens2020continuous} and grammatical structures \citep{vantrijp2012multilevel, beuls2013agent,vaneecke2018generalisation,nevens2022language,doumen2023modelling}.

\begin{figure}
\centering
\includegraphics[width = \textwidth]{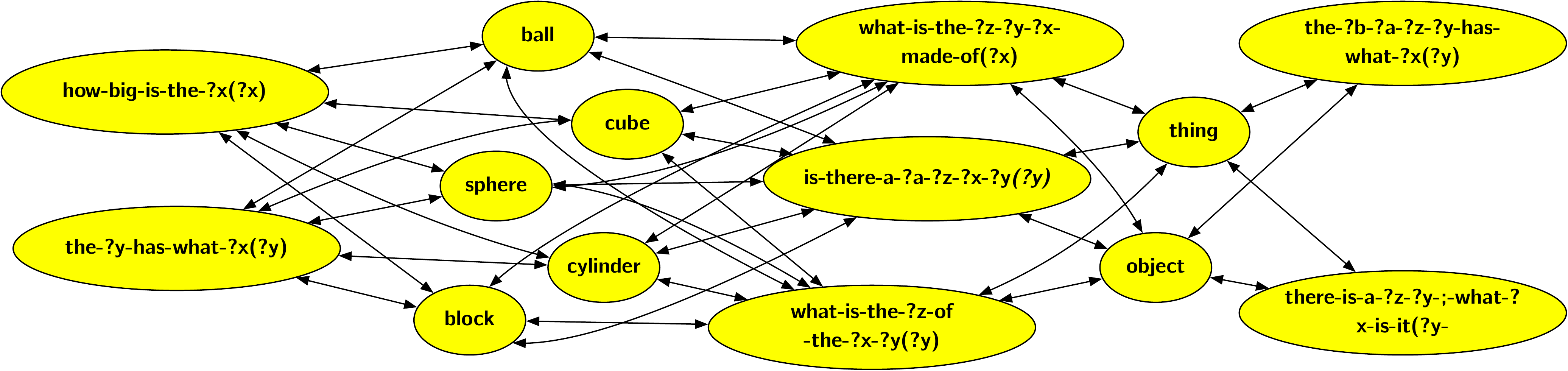}
\caption{Snapshot of a small part of an agent's categorial network built up through a question-answering game. Grammatical categories emerge as clusters within a graph that connects construction slots with their observed fillers. For example, as `ball', `cube', `sphere', `block' and `cylinder' have often been observed as fillers of the same construction slots, they are considered close in terms of grammatical category.}
\label{fig:categorial-network}
\end{figure}

\begin{figure}
\centering
\includegraphics[width = .7\textwidth]{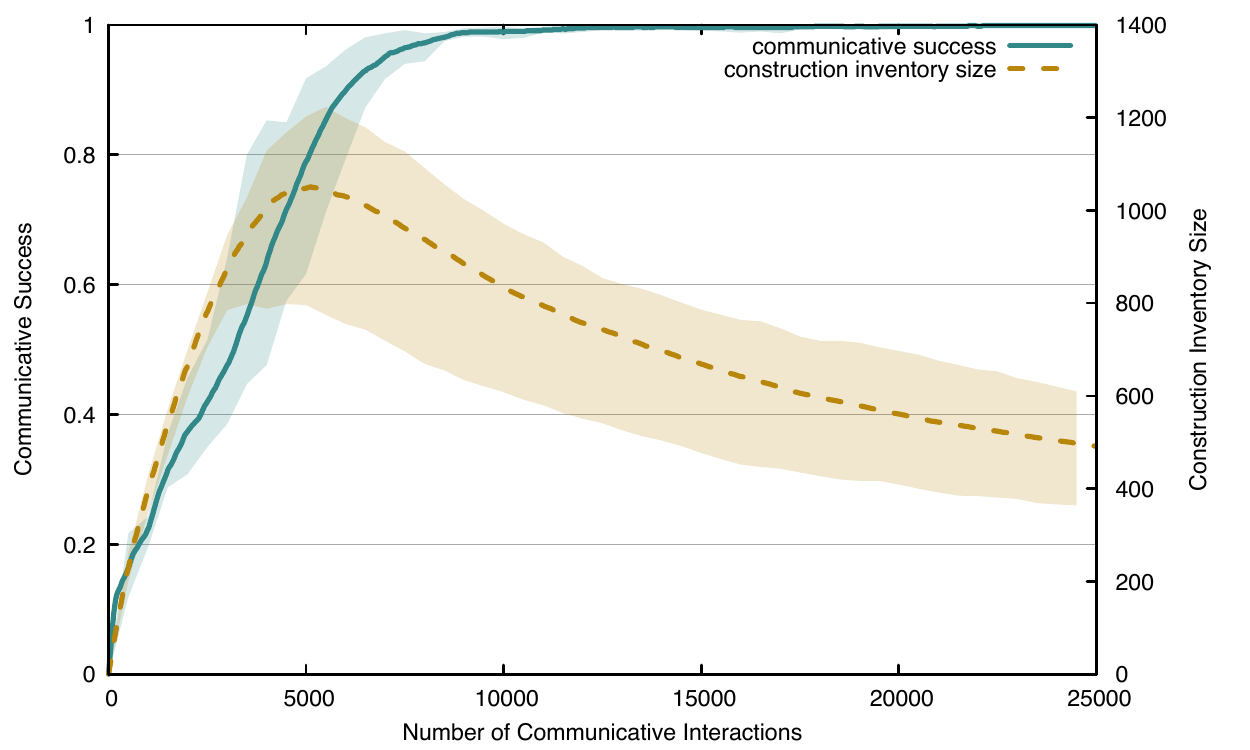}
\caption{Typical learning dynamics of a language game experiment, plotting communicative success and the construction inventory size as a function of the number of communicative interactions that have taken place. This graph has been plotted based on data from \cite{nevens2022language}.}
\label{fig:learning-dynamics}
\end{figure}

Typically, the grammatical categories that emerge during language game experiments are modelled in the form of a \textit{categorial network} with links between constructional slots and their (observed) fillers \citep{steels2022usage}, very much in the spirit of Radical Construction Grammar \citep{croft2001radical}. Categories are thus construction-specific, emergent and ever-evolving as a result of language use. Figure \ref{fig:categorial-network} shows part of a categorial network that was acquired by an artificial agent in a question-answering game \citep{nevens2022language}. We see for instance that the `ball' category is compatible with the `how-big-is-the-?x(?x)' category, reflecting the fact that when a construction has contributed a `ball' category to the transient structure, this category is compatible with the \textsc{?x} slot of the \textsc{how-big-is-the-?x-cxn}. As a consequence, grammatical categories emerge as clusters within a graph that connects construction slots with their observed fillers. For example, as `ball', `cube', `sphere', `block' and `cylinder' have often been observed as fillers of the same construction slots, they are considered close in terms of grammatical category.
A snapshot of the learning dynamics of the same language game experiment is captured in Figure \ref{fig:learning-dynamics}, where the agent's communicative success and construction inventory size are plotted as a function of time. After an initial learning phase in which the number of constructions in the learner's grammar rises steeply, the size of the grammar starts to decrease steadily as a result of the entrenchment dynamics.

\section{Construction grammar for operationalising artificial intelligence}

While the previous section has discussed the influence of insights and techniques from the field of artificial intelligence on the field of construction grammar, the present section discusses the inverse direction of influence. We will focus in particular on how the foundational ideas underlying constructionist approaches to language form an excellent fit with the needs of AI researchers who aim to build truly intelligent agents that are capable of interacting through natural language. The section starts with an overview of the desirable properties of the linguistic capability of such agents. It then continues with two use cases that illustrate the role and application of construction grammar research within the field of artificial intelligence.

\subsection{Communicatively capable intelligent agents}
\begin{figure}
\centering
\includegraphics[width =.9\textwidth]{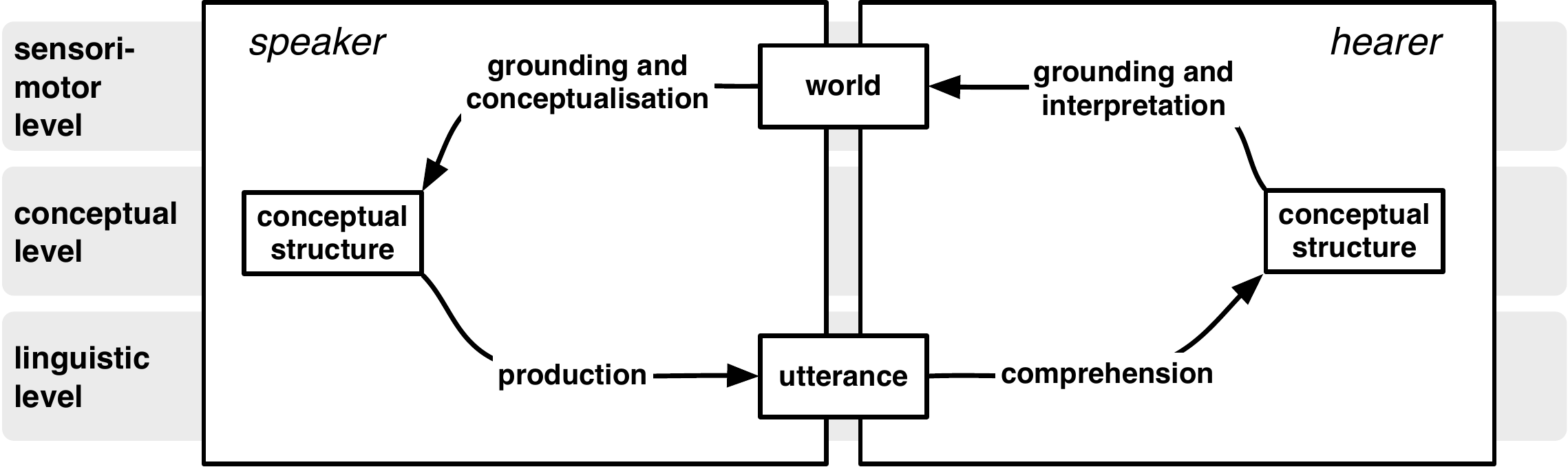}
\caption{The semiotic cycle represents the processes involved in linguistic processing from a communicative perspective. The speaker's and hearer's processes are depicted on the left-hand side and right-hand side of the figure respectively. }
\label{fig:semiotic-cycle}
\end{figure}

As introduced in the first section of this chapter, the fields of construction grammar and artificial intelligence adopt a similar attitude towards communication and language. Both fields acknowledge that language serves a communicative purpose and that language comprehension and production are equally important processes. Languages are acquired rather than innate, and they emerge and evolve as a result of communicative interactions between members of the linguistic community. Finally, languages are grounded in the world, and are therefore strongly tied to the communicative needs of their users. A graphical representation of the processes involved in language processing viewed from this perspective is shown in Figure \ref{fig:semiotic-cycle} in the form of the \textit{semiotic cycle}. The left-hand side of the semiotic cycle depicts the processes that involve the speaker and the right-hand side represents those that involve the hearer. The speaker and the hearer can both perceive the same world through their own sensors and  act upon it through their own actuators. Through a conceptualisation process, the speaker composes a conceptual structure based on its communicative intentions. In other words, the speaker decides what information it wishes to convey to the hearer and formalises this information in the form of a semantic representation. The speaker then produces an utterance that expresses this semantic representation. The hearer observes this utterance and maps it to a semantic representation of its own. This semantic representation can be seen as a reconstruction of the conceptual structure underlying the speaker's utterance based on the hearer's knowledge of the world. The hearer then interprets this conceptual structure in relation to its view on the world and acts accordingly.

Human languages are characterised by their remarkable robustness, flexibility and adaptivity to changes in the environment and communicative needs of their users. These characteristics stem from the way in which these languages have emerged and continue to evolve. The language of a community corresponds in essence to a set of conventions on which its members have converged. This global convergence is a result of purely local interactions between community members, a phenomenon often referred to as \textit{self-organisation}. At the same time, such distributed systems are inherently robust against considerable perturbations, while maintaining the flexibility to adapt to environmental changes when they occur. Construction grammar models, which start exactly from this idea and thereby incorporate the aforementioned desirable properties, are as a consequence an important source of inspiration for the field of artificial intelligence, as the same properties can be seen as crucial properties of truly intelligent agents \citep{mikolov2016roadmap}. 

\subsection{Use case 1: Modelling language acquisition in intelligent agents}

A first use case that demonstrates the application of insights and analyses from construction grammar in the field of artificial intelligence concerns the constructionist acquisition of language by intelligent agents. Usage-based constructionist theories of language acquisition argue that the ability of children to learn language is based on two general cognitive capacities: \textit{intention reading} and \textit{pattern finding} \citep{tomasello2003constructing,tomasello2009usage}. Intention reading refers to the capacity of children to understand the communicative intentions of their interlocutors, while pattern finding refers to the ability of children to recognise similarities and differences in sensori-motor experiences \cite[p. 3--4]{tomasello2003constructing}. In other terms, intention reading allows a language learner to reconstruct the meaning of an utterance that they observe during a communicative interaction, while pattern finding  provides mechanisms to learn a construction grammar based on the combination of observed utterances and their reconstructed meanings. Computational models that implement these processes are of great interest to the field of artificial intelligence, as the resulting grammars are learnable in a de-centralised, data-efficient and incremental manner.

This line of work has been embraced by a variety of researchers, pursuing goals that range from validating theories of language acquisition to finding practical solutions to problems faced by artificial agents. A first class of models operationalises pattern finding only, learning constructions from utterances paired with their meaning representation. These pairs are either provided in the form of an annotated corpus \citep{dominey2005learning,chang2008constructing,abend2017bootstrapping} or obtained through task-oriented communicative interactions in a tutor-learner scenario (see Section \ref{sec:games}) \citep{gerasymova2010acquisition,beuls2010situated,spranger2015acquisition}. A second class of models, as introduced by \cite{gaspers2011unsupervised}, is designed to learn form-meaning pairings under referential uncertainty. As such, the exact meaning representations of the input utterances are not provided to the learning algorithm, but constructions are learnt based on the combination of input utterances and situational context snippets. A third class of models operationalises both intention reading and pattern finding, whereby the results of the intention reading processes concern complex semantic structures and whereby the pattern finding processes yield constructions that generalise over pairs of observed utterances and reconstructed meaning representations \citep{nevens2022language,doumen2023modelling}.

\begin{figure}
\centering
\includegraphics[width=\textwidth]{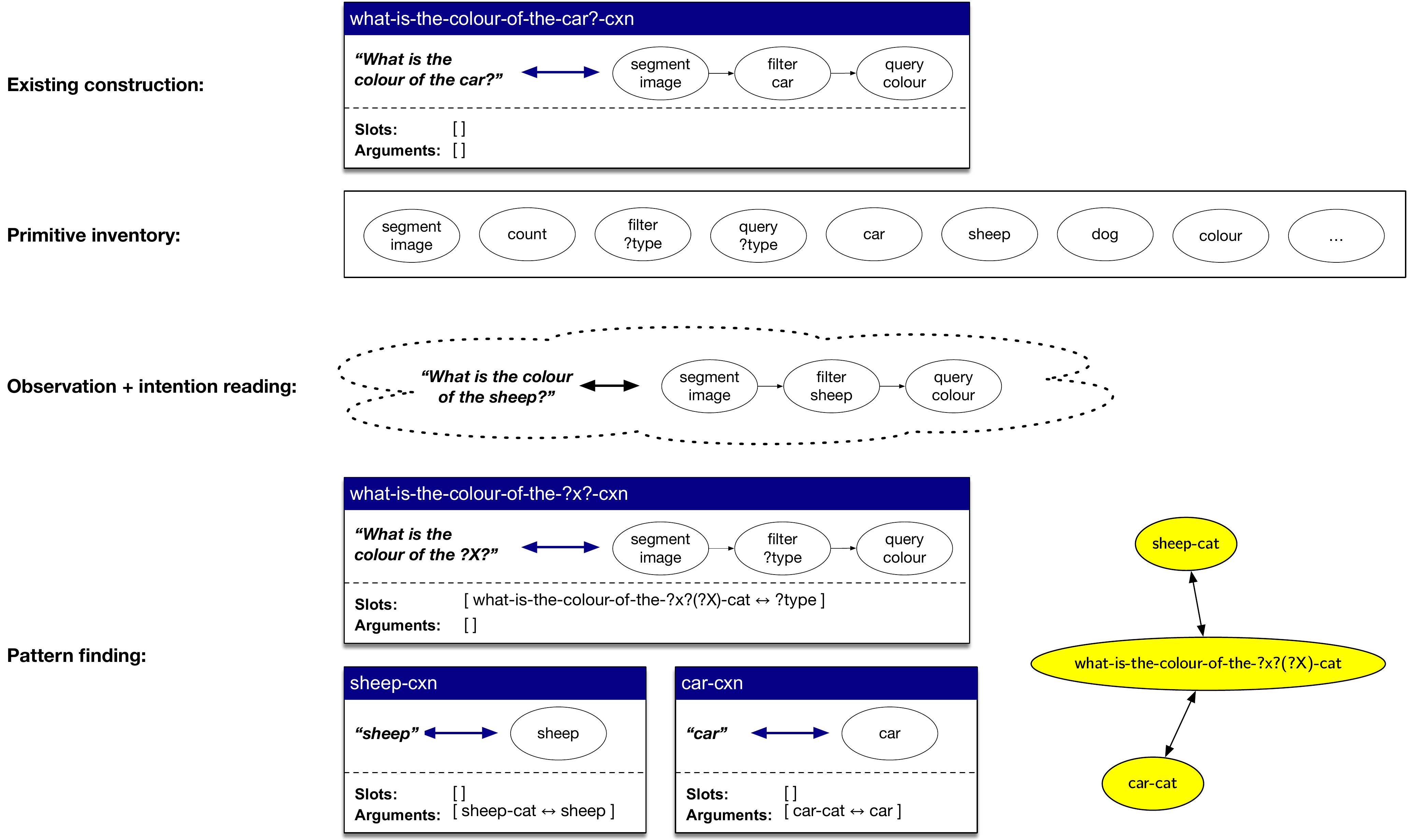}
\caption{A schematic representation of the processes of intention reading and pattern finding taking place in an artificial agent. Three new constructions and two new categorial links were learnt during a single interaction.}
\label{fig:intention-reading-pattern-finding}
\end{figure}

In general, computational models of intention reading and pattern finding operationalise task-based communicative interactions that follow the language game paradigm introduced in Section \ref{sec:games}. They thereby implement the processes of grounding, conceptualisation and interpretation, and language comprehension and production depicted in the semiotic cycle in Figure \ref{fig:semiotic-cycle}.  Imagine that an agent needs to learn to answer questions about the world it visually observes. At the beginning of the experiment, the agent only knows how to perform a limited number of cognitive operations. These operations include for example segmenting an image, filtering a set according to a prototype, counting the number of items in a set, and querying properties of an object. The agent does not know any constructions or other linguistic entities such as grammatical categories or word boundaries at the start of the experiment. A tutor agent might ask the learner agent to name the colour of the car that passes by using the utterance ``\textit{What is the colour of the car?}''. At this point, the learner agent will signal that it does not understand the utterance and the tutor agent will provide the answer to the question as feedback (e.g. \textsc{yellow}). Based on this answer, the learner agent will make a hypothesis about the intended meaning of the observed utterance. In order to do this, it will compose a semantic network based on the primitive cognitive operations it knows, such that, upon evaluation, this network leads to the answer that was provided by the tutor (e.g. [segment image - filter car - query colour]). If later, the tutor asks ``\textit{What is the colour of the sheep?}'' and the learner hypothesises after feedback that it means [segment image - filter sheep - query colour], the learner can construct a generalised pattern that pairs ``\textit{What is the colour of the ?X}'' with the meaning representation [segment image - filter ?X - query colour]. At the same time, the learner can learn two patterns which pair ``sheep'' and ``car'' with their respective meanings,  as well as two links in its categorial network that express that `sheep' and `car' can fill the `?X' slot in the \textsc{what-is-the-colour-of-the-?x-cxn}. A schematic representation of the intention reading and pattern finding processes involved in the processing of this example is shown in Figure \ref{fig:intention-reading-pattern-finding}.

The entrenchment dynamics of the game ensure that after many interactions, the linguistic model of the learner agent is compatible with the tutor's language use in their shared environment. The composition of semantic networks as hypotheses about the intended meaning of the other agent constitutes an operationalisation of intention reading, while the syntactico-semantic generalisation over pairs of utterances and semantic networks constitutes an operationalisation of pattern finding. 

The constructionist acquisition of language through intention reading and pattern finding in task-based communicative interactions constitutes a paradigm that combines a number of features that are highly valued in the field of artificial intelligence. The paradigm assumes that agents are autonomous entities which sense, reason and act independently. The global behaviour that arises in the community stems from purely local interactions and is robust and adaptive as a consequence of its evolutionary nature. The paradigm focusses on the meaning and intentions underlying language as well as on their grounding in both the world and the knowledge of the agents. Semantic structures are composed by the agents themselves based on the environment, communicative feedback and mental simulation. Learning is data-efficient and problem-driven, with one-shot learning of constructions being the norm. As the constructions that result from the learning process can generalise over the compositional aspects of the language (both in terms of form and meaning) and keep the non-compositional aspects within the constructions, the paradigm is compatible with any meaning representation. It is perhaps this insight from construction grammar, namely that constructions can elegantly handle non-compositional forms and meanings, that has led to its appreciation in the artificial intelligence community.  The agents in the population do not even need to share the same primitive operations, morphology or software architecture, making it possible to have communicatively adequate languages emerge in populations of heterogeneous agents.
 Finally, both the learning process and the resulting grammars are fully explainable and human interpretable, which excellently fits the current focus on explainable and trustworthy AI.

\subsection{Use case 2: Modelling opinion dynamics for understanding society}

The second use case that showcases the application potential of construction grammar within the field of artificial intelligence concerns the automatic analysis of opinions expressed on social media platforms. Today, such platforms play an important role in the formation of opinions and their propagation throughout society. The enormous amounts of data created every day make it impossible to grasp the dynamic landscape of opinions held by the members of a community. Automatic analysis tools therefore play an important role as research instruments for social scientists investigating this matter. Such tools need to be capable of analysing social media posts and situate the opinions they express with respect to other opinions as well as real-world events. An important aspect of these tools is their ability to reason over the meaning of textual documents. Large-scale construction grammars can play an important role in the semantic analysis of these documents, as they are able to retrieve their underlying meaning through a transparent and interpretable model. 

An illustrative example of an application that makes use of construction grammar to analyse opinion dynamics is the Penelope  opinion facilitator \citep{willaert2020building,willaert2021opinion}. The opinion facilitator aims to help consumers of online news media to investigate on the fly statements made in an article or newspaper comment by presenting other articles or comments that put forward concurring or diverging opinions. The news consumer is thereby offered a broad spectrum of opinions about a subject matter, reducing the risk of getting drawn into an echo chamber \citep[see e.g.][]{sunstein2018republic}.

\begin{figure}
\centering
\includegraphics[width=\textwidth]{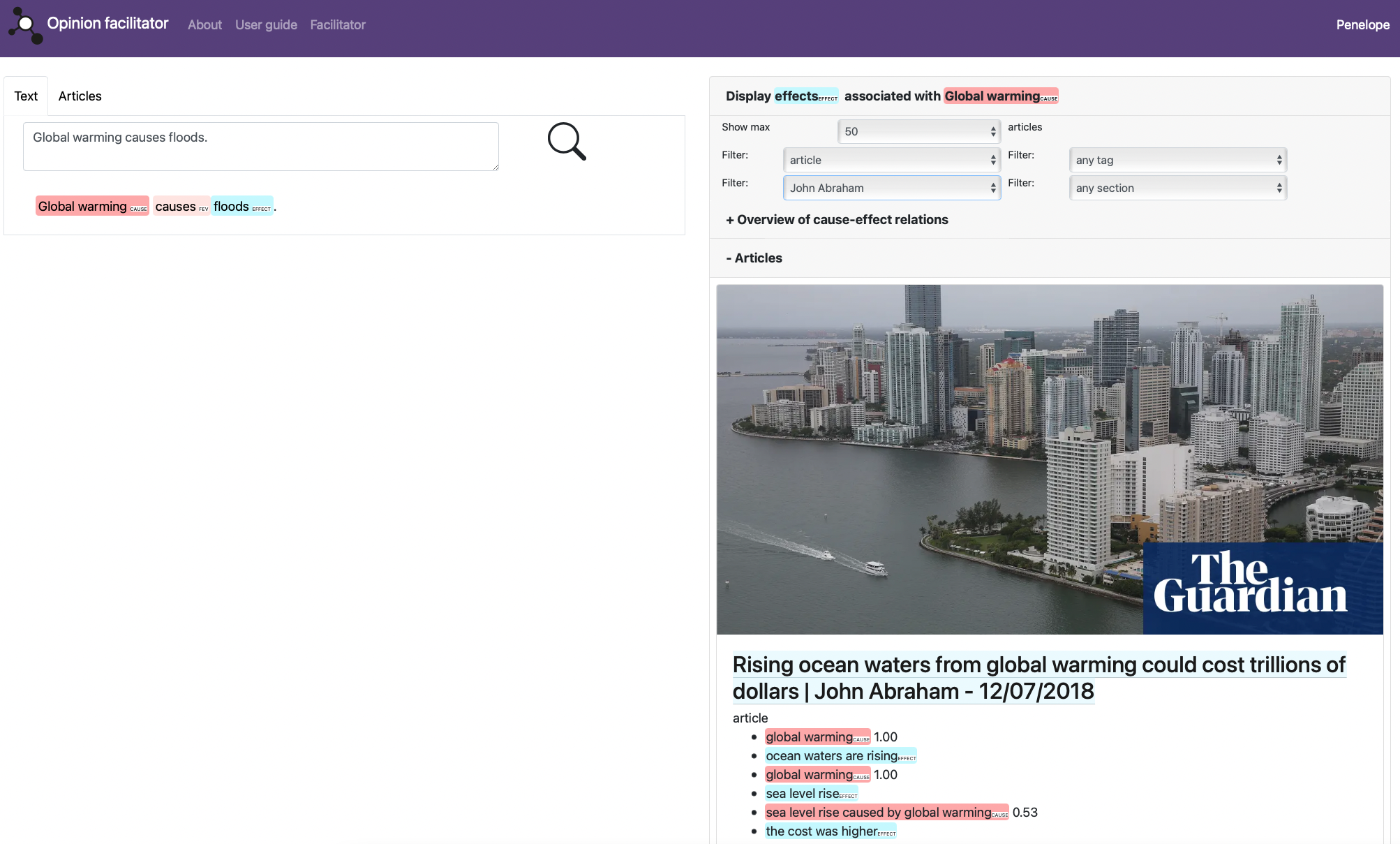}
\caption{The Penelope opinion facilitator \citep{willaert2021opinion} makes use of a computational construction grammar to provide a frame-semantic analysis of statements, news articles and comments. Based on this analysis, articles and comments expressing concurrent and diverging opinions are revealed.}
\label{fig:opinion-facilitator}
\end{figure}

An example of the use of the opinion facilitator is shown in Figure \ref{fig:opinion-facilitator}. At the left-hand side of the interface, the user can either enter a statement they wish to investigate or browse newspaper articles. A frame-semantic analysis of the statement or article is then visualised. In this visualisation, the user can click on a participant role of a frame. Articles that contain the same frame with a semantically similar filler for the role that was clicked are then shown on the right-hand side of the interface. In these articles, the relevant frames are highlighted and a short summary is provided. In the example in Figure \ref{fig:opinion-facilitator}, the user has entered the statement ``\textit{Global warming causes floods}''. A causal frame was detected, with ``\textit{global warming}'' filling the cause slot and ``\textit{floods}'' filling the effect slot. The user has clicked on ``\textit{global warming}'' and articles containing causal frames with global warming as cause are displayed on the right. The user has thereby found a broad spectrum of articles that mention effects of global warming. The user can then build an informed opinion about the original statement based on the information conveyed through these articles. In this application, the frame-semantic analysis of the texts is performed by a computational construction grammar \citep{beuls2021computational}. The main advantage of the use of such a grammar is that it is entirely transparent and human-interpretable. The detection of a frame and the assignment of participant roles is always the consequence of a construction application, and is thereby  linguistically motivated and explainable.

\section{Discussion and conclusion}

Throughout this chapter, our aim has been to draw renewed attention to the common ground that is shared by the fields of construction grammar and artificial intelligence. We have done this on the one hand through a discussion of their historical ties and their common attitude towards communication and language, and on the other hand through an analysis of the way in which both fields have influenced and continue to influence each other. 

When the field of construction grammar emerged in the eighties, its common ground with the field of artificial intelligence was evident to its architects. Indeed, it was clear that both fields shared the objective of modelling human communication and language, and that they held a similar attitude towards the nature of the subject matter. Both fields acknowledge that language serves as an instrument of communication between members of a community and that it has emerged and continues to evolve to serve its communicative purpose. As a logical consequence, both fields emphasise the importance of modelling language use, including the processes of language comprehension and production, rather than studying the competence of an ideal language user. Both fields acknowledge that languages are acquired rather than innate, and that they emerge and evolve as a consequence of local communicative interactions between community members. The linguistic system of each community member is therefore unique as it has been shaped by their past successes and failures in communication. Finally,  linguistic systems are grounded in the environment and world knowledge of community members, and are adaptive to changes in the environment and communicative needs of their users. We can only speculate about the reasons that the knowledge of this common ground has faded away in both research communities, but are convinced that a renewed awareness of their shared objectives and attitude towards communication and language will benefit future research in both fields.

When it comes to the first direction of influence, i.e. the influence of the field of artificial intelligence on the field of construction grammar, we have argued that ideas and techniques from AI have played a crucial role in the formalisation and computational implementation of the basic tenets of construction grammar. A wide range of AI techniques has been deployed in this endeavour. Feature structures are used to formalise constructions and innovative unification algorithms have been developed to operationalise the processes of construction-based language comprehension and production. Constructional language processing is operationalised as problem solving through search, with heuristic search strategies making the free combination of constructions computationally tractable. Finally, multi-agent simulations are used to model the emergence, acquisition and dynamic evolution of grounded construction grammars within populations of language users. In sum, insights and techniques from the field of artificial intelligence have served as a cornerstone in the operationalisation of computational construction grammar.

In regard to the second direction of influence, i.e. the influence of the field of construction grammar on the field of artificial intelligence, we have highlighted that the foundational ideas underlying construction grammar form an excellent fit with the artificial intelligence goal of building communicatively capable agents. First of all, the focus on the meaning of linguistic expressions, rather than on their form, and the grounding of this meaning in the world and knowledge of language users, supports the development of AI systems that can interact with their environment and each other through natural language. Second, the dynamic and usage-based nature of constructions, combined with the de-centralised nature of their acquisition, facilitates the bootstrapping of communication systems that exhibit the robustness, flexibility and adaptivity found in human languages. Finally, the inherent ability of constructions to generalise over the compositional aspects of language use and to capture any aspects of language use where the form and meaning are non-compositional with respect to each other is perhaps the most desirable property of construction grammar when it comes to building real-world AI systems. We have illustrated these aspects through two use cases. The first use case concerned the modelling of the acquisition of a construction grammar that enables  an autonomous agent to learn to answer questions about its environment. The second use case presented an opinion facilitator tool in which frame-semantic analyses obtained through a human-interpretable computational construction grammar served as the basis for tracking opinions in on-line news media.

We strongly believe that a revaluation and further elaboration of the strong relationship between the research fields of construction grammar and artificial intelligence will play a key role in shaping the future of the field of construction grammar. Indeed, computational operationalisations of construction grammar  bring important methodological advantages that bear the promise of leading to a number of substantial breakthroughs with respect to the state of the art. Most prominently, computational operationalisations are indispensable when it comes to scaling constructionist approaches to language. They facilitate the automatic validation of the preciseness and internal consistency of construction grammar theories and analyses, which is impossible to do by hand for grammars that consist of tens of thousands of constructions. Moreover, they allow to corroborate constructionist analyses with large amounts of corpus data, unequivocally revealing what they can and cannot account for. The scalability advantages of computational construction grammar also support moving away from the study of individual constructions to the study of systemic relations between families of constructions, thereby directly contributing to theory formation. An additional benefit of computational operationalisations concerns the standardisation of the way in which constructions are represented, thereby facilitating the exchange of ideas and results among researchers. Finally, computational operationalisations will play a crucial role in enhancing the application potential of construction grammar, both as a linguistic framework adopted in a variety of  other scientific disciplines and as a central component of communicatively capable AI systems.

\FloatBarrier


\newpage
\begin{thebibliography}{}

\bibitem[Abend et~al., 2017]{abend2017bootstrapping}
Abend, O., Kwiatkowski, T., Smith, N.~J., Goldwater, S., and Steedman, M.
  (2017).
\newblock Bootstrapping language acquisition.
\newblock {\em Cognition}, 164:116--143.

\bibitem[Baker et~al., 1998]{baker1998berkeley}
Baker, C., Fillmore, C., and Lowe, J. (1998).
\newblock The berkeley framenet project.
\newblock In {\em Proceedings of the 17th international conference on
  Computational linguistics-Volume 1}, pages 86--90. Association for
  Computational Linguistics.

\bibitem[Banarescu et~al., 2013]{banarescu2013abstract}
Banarescu, L., Bonial, C., Cai, S., Georgescu, M., Griffitt, K., Hermjakob, U.,
  Knight, K., Koehn, P., Palmer, M., and Schneider, N. (2013).
\newblock Abstract meaning representation for sembanking.
\newblock In {\em Proceedings of the 7th Linguistic Annotation Workshop and
  Interoperability with Discourse}, pages 178--186.

\bibitem[Baronchelli et~al., 2006]{baronchelli2006sharp}
Baronchelli, A., Felici, M., Loreto, V., Caglioti, E., and Steels, L. (2006).
\newblock Sharp transition towards shared vocabularies in multi-agent systems.
\newblock {\em Journal of Statistical Mechanics: Theory and Experiment},
  2006(06):P06014.

\bibitem[Bergen and Chang, 2005]{bergen2005embodied}
Bergen, B. and Chang, N. (2005).
\newblock {Embodied Construction Grammar} in simulation-based language
  understanding.
\newblock In Fried, M. and {\"O}stman, J.-O., editors, {\em {Construction
  Grammars}: Cognitive Grounding and Theoretical Extensions}, pages 147--190.
  John Benjamins, Amsterdam, Netherlands.

\bibitem[Beuls et~al., 2010]{beuls2010situated}
Beuls, K., Gerasymova, K., and {van Trijp}, R. (2010).
\newblock Situated learning through the use of language games.
\newblock In {\em {Proceedings of the 19th Annual Machine Learning Conference
  of Belgium and The Netherlands (BeNeLearn)}}, pages 1--6.

\bibitem[Beuls and Steels, 2013]{beuls2013agent}
Beuls, K. and Steels, L. (2013).
\newblock Agent-based models of strategies for the emergence and evolution of
  grammatical agreement.
\newblock {\em PLOS ONE}, 8(3):e58960.

\bibitem[Beuls and {Van Eecke}, 2023]{beuls2023fluid}
Beuls, K. and {Van Eecke}, P. (2023).
\newblock {Fluid Construction Grammar}: State of the art and future outlook.
\newblock In Bonial, C. and {Tayyar Madabushi}, H., editors, {\em Proceedings
  of the First International Workshop on Construction Grammars and {NLP}
  ({CxGs+NLP}, {GURT/SyntaxFest 2023})}, pages 41--50.

\bibitem[Beuls et~al., 2021]{beuls2021computational}
Beuls, K., {Van Eecke}, P., and Cangalovic, V.~S. (2021).
\newblock A computational construction grammar approach to semantic frame
  extraction.
\newblock {\em Linguistics Vanguard}, 7(1):20180015.

\bibitem[Bleys et~al., 2011]{bleys2011search}
Bleys, J., Stadler, K., and {De Beule}, J. (2011).
\newblock Search in linguistic processing.
\newblock In Steels, L., editor, {\em Design Patterns in Fluid Construction
  Grammar}, pages 149--179. John Benjamins, Amsterdam, Netherlands.

\bibitem[Bleys and Steels, 2009]{bleys2009linguistic}
Bleys, J. and Steels, L. (2009).
\newblock Linguistic selection of language strategies, a case study for color.
\newblock In {\em {Proceedings of the 10th {European} Conference on Artifical
  Life}}, pages 150--157.

\bibitem[Bonial et~al., 2018]{bonial2018abstract}
Bonial, C., Badarau, B., Griffitt, K., Hermjakob, U., Knight, K., O{'}Gorman,
  T., Palmer, M., and Schneider, N. (2018).
\newblock {A}bstract {M}eaning {R}epresentation of constructions.
\newblock In {\em Proceedings of the Eleventh International Conference on
  Language Resources and Evaluation ({LREC} 2018)}. European Language Resources
  Association (ELRA).

\bibitem[Chang, 2008]{chang2008constructing}
Chang, N. (2008).
\newblock {\em Constructing grammar: A computational model of the emergence of
  early constructions}.
\newblock PhD thesis, University of California, Berkeley, Berkeley, CA, USA.

\bibitem[Croft, 2001]{croft2001radical}
Croft, W. (2001).
\newblock {\em Radical construction grammar: Syntactic theory in typological
  perspective}.
\newblock Oxford University Press, Oxford, United Kingdom.

\bibitem[{de Boer}, 2000]{deboer2000self}
{de Boer}, B. (2000).
\newblock Self-organization in vowel systems.
\newblock {\em Journal of Phonetics}, 28(4):441--465.

\bibitem[{De Vylder} and Tuyls, 2006]{devylder2006reach}
{De Vylder}, B. and Tuyls, K. (2006).
\newblock How to reach linguistic consensus.
\newblock {\em Journal of Theoretical Biology}, 242(4):818--831.

\bibitem[Dominey and Boucher, 2005]{dominey2005learning}
Dominey, P.~F. and Boucher, J.-D. (2005).
\newblock Learning to talk about events from narrated video in a construction
  grammar framework.
\newblock {\em Artificial Intelligence}, 167(1):31--61.

\bibitem[Doumen et~al., 2023]{doumen2023modelling}
Doumen, J., Beuls, K., and {Van Eecke}, P. (2023).
\newblock Modelling language acquisition through syntactico-semantic pattern
  finding.
\newblock In Vlachos, A. and Augenstein, I., editors, {\em Findings of the
  {Association for Computational Linguistics}}, pages 1317--1327. Association
  for Computational Linguistics.

\bibitem[Feldman et~al., 2009]{feldman2009embodied}
Feldman, J., Dodge, E., and Bryant, J. (2009).
\newblock Embodied construction grammar.
\newblock In Heine, B. and Narrog, H., editors, {\em The {Oxford} Handbook of
  Linguistic Analysis}, pages 121--146. Oxford University Press, Oxford, United
  Kingdom.

\bibitem[Fillmore, 1968]{fillmore1968case}
Fillmore, C. (1968).
\newblock The case for case.
\newblock In Bach, E.~W. and Harms, R.~T., editors, {\em Universals in
  Linguistic Theory}, pages 1--88. Holt, Rinehart \& Winston, New York, NY,
  USA.

\bibitem[Fillmore et~al., 1988]{fillmore1988regularity}
Fillmore, C., Kay, P., and O'connor, M. (1988).
\newblock Regularity and idiomaticity in grammatical constructions.
\newblock {\em Language}, 64(3):501--538.

\bibitem[Fillmore, 1976]{fillmore1976frame}
Fillmore, C.~J. (1976).
\newblock Frame semantics and the nature of language.
\newblock In {\em Annals of the New York Academy of Sciences: Conference on the
  origin and development of language and speech}, volume 280, pages 20--32. New
  York, NY, USA.

\bibitem[Fillmore, 1988]{fillmore1988mechanisms}
Fillmore, C.~J. (1988).
\newblock The mechanisms of ``construction grammar''.
\newblock In {\em Annual Meeting of the Berkeley Linguistics Society},
  volume~14, pages 35--55.

\bibitem[Fillmore and Baker, 2001]{fillmore2001frame}
Fillmore, C.~J. and Baker, C.~F. (2001).
\newblock Frame semantics for text understanding.
\newblock In {\em Proceedings of WordNet and Other Lexical Resources Workshop,
  NAACL}, volume~6.

\bibitem[Fried and {\"O}stman, 2004]{fried2004construction}
Fried, M. and {\"O}stman, J.-O. (2004).
\newblock Construction grammar: A thumbnail sketch.
\newblock In {\"O}stman, J.-O. and Fried, M., editors, {\em Construction
  grammar in a cross-language perspective}, pages 1--86. John Benjamins,
  Amsterdam, Netherlands.

\bibitem[Gaspers et~al., 2011]{gaspers2011unsupervised}
Gaspers, J., Cimiano, P., Griffiths, S.~S., and Wrede, B. (2011).
\newblock An unsupervised algorithm for the induction of constructions.
\newblock In {\em Proceedings of the 2011 {IEEE} International Conference on
  Development and Learning (ICDL)}, volume~2, pages 1--6. IEEE.

\bibitem[Gerasymova and Spranger, 2010]{gerasymova2010acquisition}
Gerasymova, K. and Spranger, M. (2010).
\newblock Acquisition of grammar in autonomous artificial systems.
\newblock In Coelho, M., Studer, R., and Woolridge, M., editors, {\em
  Proceedings of the 19th {European} Conference on Artificial Intelligence
  (ECAI-2010)}, pages 923--928.

\bibitem[Goffman, 1974]{goffman1974frame}
Goffman, E. (1974).
\newblock {\em Frame analysis: An essay on the organization of experience.}
\newblock Harvard University Press.

\bibitem[Goldberg, 1995]{goldberg1995constructions}
Goldberg, A.~E. (1995).
\newblock {\em Constructions: A construction grammar approach to argument
  structure}.
\newblock University of Chicago Press, Chicago, IL, USA.

\bibitem[Goldberg, 2006]{goldberg2006constructions}
Goldberg, A.~E. (2006).
\newblock {\em Constructions at work}.
\newblock Oxford University Press, Oxford, United Kingdom.

\bibitem[Hilpert, 2014]{hilpert2014construction}
Hilpert, M. (2014).
\newblock {\em Construction grammar and its application to {English}}.
\newblock Edinburgh University Press, Edinburgh, United Kingdom.

\bibitem[Hilpert, 2021]{hilpert2021what}
Hilpert, M. (2021).
\newblock {\em What is construction grammar?}, pages 1--35.
\newblock Distinguished lectures in cognitive linguistics. Brill, Leiden, The
  Netherlands.

\bibitem[Jurafsky, 2014]{jurafsky2014charles}
Jurafsky, D. (2014).
\newblock Charles j. fillmore.
\newblock {\em Computational Linguistics}, 40(3):725--731.

\bibitem[Kay and Fillmore, 1999]{kay1999grammatical}
Kay, P. and Fillmore, C. (1999).
\newblock Grammatical constructions and linguistic generalizations.
\newblock {\em Language}, 75(1):1--33.

\bibitem[Michaelis, 2013]{michaelis2013sign}
Michaelis, L.~A. (2013).
\newblock Sign-based construction grammar.
\newblock {\em The Oxford handbook of construction grammar}, pages 133--152.

\bibitem[Mikolov et~al., 2016]{mikolov2016roadmap}
Mikolov, T., Joulin, A., and Baroni, M. (2016).
\newblock A roadmap towards machine intelligence.
\newblock In {\em Proceedings of the International Conference on Intelligent
  Text Processing and Computational Linguistics}, pages 29--61.

\bibitem[Minsky, 1974]{minsky1974framework}
Minsky, M. (1974).
\newblock A framework for representing knowledge.
\newblock Technical report, Technical Report 306, MIT AI Laboratory, Cambridge,
  MA, USA.

\bibitem[Nevens et~al., 2022]{nevens2022language}
Nevens, J., Doumen, J., {Van Eecke}, P., and Beuls, K. (2022).
\newblock Language acquisition through intention reading and pattern finding.
\newblock In Calzolari, N. and Huang, C.-R., editors, {\em Proceedings of the
  29th International Conference on Computational Linguistics}, pages 15--25.
  International Committee on Computational Linguistics.

\bibitem[Nevens et~al., 2020]{nevens2020continuous}
Nevens, J., {Van Eecke}, P., and Beuls, K. (2020).
\newblock From continuous observations to symbolic concepts: A
  discrimination-based strategy for grounded concept learning.
\newblock {\em Frontiers in Robotics and AI}, 7(84).

\bibitem[Newell and Simon, 1956]{newell1956logic}
Newell, A. and Simon, H. (1956).
\newblock The logic theory machine – a complex information processing system.
\newblock {\em IRE Transactions on Information Theory}, 2(3):61--79.

\bibitem[Oudeyer, 2006]{oudeyer2006self}
Oudeyer, P.-Y. (2006).
\newblock {\em Self-organization in the evolution of speech}.
\newblock Oxford University Press, Oxford, United Kingdom.

\bibitem[Pearl, 1984]{pearl1984heuristics}
Pearl, J. (1984).
\newblock {\em Heuristics}.
\newblock Addison-Wesley Longman Publishing Co., Inc.

\bibitem[Rumelhart, 1980]{rumelhart1080schemata}
Rumelhart, D.~E. (1980).
\newblock Schemata: The building blocks of cognition.
\newblock In Spiro, R., Bruce, B., and Brewer, W., editors, {\em Theoretical
  Issues in Reading Comprehension}, pages 33--58. Lawrence Erlbaum Associates,
  Hillsdale, NJ, USA.

\bibitem[Russell and Norvig, 2021]{russell2021artificial}
Russell, S. and Norvig, P. (2021).
\newblock {\em Artificial intelligence}.
\newblock Pearson, Hoboken, NJ, USA, 4th edition.

\bibitem[Sag, 2012]{sag2012sign}
Sag, I.~A. (2012).
\newblock Sign-based construction grammar: An informal synopsis.
\newblock In Boas, H.~C. and Sag, I.~A., editors, {\em Sign-based construction
  grammar}, pages 69--202. CSLI Publications/Center for the Study of Language
  and Information, Stanford, CA, USA.

\bibitem[Schank and Abelson, 1977]{schank1977scripts}
Schank, R.~C. and Abelson, R.~P. (1977).
\newblock {\em Scripts, plans, goals, and understanding: An inquiry into human
  knowledge structures}.
\newblock Lawrence Erlbaum Associates, Hillsdale, NJ, USA.

\bibitem[{Sierra Santib{\'a}{\~n}ez}, 2012]{sierra2012logic}
{Sierra Santib{\'a}{\~n}ez}, J. (2012).
\newblock A logic programming approach to parsing and production in {Fluid
  Construction Grammar}.
\newblock In Steels, L., editor, {\em Computational Issues in {Fluid
  Construction Grammar}}, volume 7249 of {\em Lecture Notes in Computer
  Science}, pages 239--255. Springer, Berlin, Germany.

\bibitem[Spranger, 2016]{spranger2016evolution}
Spranger, M. (2016).
\newblock {\em The evolution of grounded spatial language}.
\newblock Language Science Press, Berlin, Germany.

\bibitem[Spranger and Steels, 2015]{spranger2015acquisition}
Spranger, M. and Steels, L. (2015).
\newblock Co-acquisition of syntax and semantics: an investigation in spatial
  language.
\newblock In Yang, Q. and Wooldridge, M., editors, {\em Proceedings of the
  Twenty-Fourth International Joint Conference on Artificial Intelligence},
  pages 1909--1915, Palo Alto, CA, USA. AAAI Press.

\bibitem[Steels, 1998]{steels1998origins}
Steels, L. (1998).
\newblock The origins of syntax in visually grounded robotic agents.
\newblock {\em Artificial Intelligence}, 103(1-2):133--156.

\bibitem[Steels, 2001]{steels2001language}
Steels, L. (2001).
\newblock Language games for autonomous robots.
\newblock {\em {IEEE} Intelligent Systems}, 16:16--22.

\bibitem[Steels, 2004]{steels2004constructivist}
Steels, L. (2004).
\newblock Constructivist development of grounded construction grammar.
\newblock In {\em Proceedings of the 42nd Annual Meeting of the {Association
  for Computational Linguistics} ({ACL}-04)}, pages 9--16.

\bibitem[Steels, 2005]{steels2005emergence}
Steels, L. (2005).
\newblock The emergence and evolution of linguistic structure.
\newblock {\em Connection Science}, 17:213--230.

\bibitem[Steels, 2015]{steels2015talking}
Steels, L. (2015).
\newblock {\em The Talking Heads experiment}.
\newblock Language Science Press, Berlin, Germany.

\bibitem[Steels and Belpaeme, 2005]{steels2005coordinating}
Steels, L. and Belpaeme, T. (2005).
\newblock Coordinating perceptually grounded categories through language.
\newblock {\em Behavioral and Brain Sciences}, 28(4):469--488.

\bibitem[Steels and {De Beule}, 2006]{debeule2006unify}
Steels, L. and {De Beule}, J. (2006).
\newblock Unify and merge in fluid construction grammar.
\newblock In Vogt, P., Sugita, Y., Tuci, E., and Nehaniv, C., editors, {\em
  Symbol Grounding and Beyond}, pages 197--223, Heidelberg, Germany. Springer.

\bibitem[Steels et~al., 2022]{steels2022usage}
Steels, L., {Van Eecke}, P., and Beuls, K. (2022).
\newblock Usage-based learning of grammatical categories.
\newblock {\em arXiv preprint arXiv:2204.10201}.

\bibitem[Sunstein, 2018]{sunstein2018republic}
Sunstein, C.~R. (2018).
\newblock {\em \# Republic}.
\newblock Princeton University Press, Princeton, NJ, USA.

\bibitem[Tomasello, 2003]{tomasello2003constructing}
Tomasello, M. (2003).
\newblock {\em Constructing a Language: A Usage-Based Theory of Language
  Acquisition}.
\newblock Harvard University Press, Harvard, MA, USA.

\bibitem[Tomasello, 2009]{tomasello2009usage}
Tomasello, M. (2009).
\newblock The usage-based theory of language acquisition.
\newblock In {\em The {Cambridge} handbook of child language}, pages 69--87.
  Cambridge University Press, Cambridge, United Kingdom.

\bibitem[{Van Eecke}, 2018]{vaneecke2018generalisation}
{Van Eecke}, P. (2018).
\newblock {\em Generalisation and specialisation operators for computational
  construction grammar and their application in evolutionary linguistics
  Research}.
\newblock PhD thesis, Vrije Universiteit Brussel, Brussels: VUB Press.

\bibitem[{Van Eecke} and Beuls, 2017]{vaneecke2017metalayer}
{Van Eecke}, P. and Beuls, K. (2017).
\newblock Meta-layer problem solving for computational construction grammar.
\newblock In {\em The 2017 {AAAI Spring} Symposium Series}, pages 258--265.
  AAAI Press.

\bibitem[{Van Eecke} and Beuls, 2018]{vaneecke2018creative}
{Van Eecke}, P. and Beuls, K. (2018).
\newblock Exploring the creative potential of computational construction
  grammar.
\newblock {\em Zeitschrift f\"ur Anglistik und Amerikanistik}, 66(3):341--355.

\bibitem[{Van Eecke} et~al., 2022]{vaneecke2022neural}
{Van Eecke}, P., Nevens, J., and Beuls, K. (2022).
\newblock Neural heuristics for scaling constructional language processing.
\newblock {\em Journal of Language Modelling}, 10(2):287--314.

\bibitem[{van Trijp}, 2008]{vantrijp2008emergence}
{van Trijp}, R. (2008).
\newblock The emergence of semantic roles in fluid construction grammar.
\newblock In {\em The Evolution Of Language}, pages 346--353. World Scientific.

\bibitem[{van Trijp}, 2016]{vantrijp2016chopping}
{van Trijp}, R. (2016).
\newblock Chopping down the syntax tree.
\newblock {\em Belgian Journal of Linguistics}, 30(1):15--38.

\bibitem[{van Trijp} et~al., 2022]{vantrijp2022fcg}
{van Trijp}, R., Beuls, K., and {Van Eecke}, P. (2022).
\newblock The {FCG} {E}ditor: An innovative environment for engineering
  computational construction grammars.
\newblock {\em PLOS ONE}, 17(6):e0269708.

\bibitem[{van Trijp} and Steels, 2012]{vantrijp2012multilevel}
{van Trijp}, R. and Steels, L. (2012).
\newblock Multilevel alignment maintains language systematicity.
\newblock {\em Advances in Complex Systems}, 15(3--4):1250039.

\bibitem[Wellens and {De Beule}, 2010]{wellens2010priming}
Wellens, P. and {De Beule}, J. (2010).
\newblock Priming through constructional dependencies: a case study in {Fluid
  Construction Grammar}.
\newblock In {\em The Evolution of Language: Proceedings of the 8th
  International Conference (EVOLANG8)}, pages 344--351. World Scientific.

\bibitem[Willaert et~al., 2020]{willaert2020building}
Willaert, T., {Van Eecke}, P., Beuls, K., and Steels, L. (2020).
\newblock Building social media observatories for monitoring online opinion
  dynamics.
\newblock {\em Social Media + Society}, 6(2).

\bibitem[Willaert et~al., 2021]{willaert2021opinion}
Willaert, T., {Van Eecke}, P., {Van Soest}, J., and Beuls, K. (2021).
\newblock An opinion facilitator for online news media.
\newblock {\em Frontiers in Big Data}, 4:1--10.

\bibitem[Winston, 1977]{winston1977artificial}
Winston, P.~H. (1977).
\newblock {\em Artificial Intelligence}.
\newblock Addison Wesley, Reading, MA, USA.

\end{thebibliography}
\end{document}